\documentclass[letterpaper, 10 pt, conference]{ieeeconf}  

\IEEEoverridecommandlockouts                              

\usepackage{graphics} 
\usepackage{epsfig} 
\usepackage{mathptmx} 
\usepackage{times} 
\usepackage{amsmath} 
\usepackage{amssymb}  
\usepackage{amsfonts}
\usepackage{mathrsfs}
\usepackage{algorithmic}
\usepackage{algorithm}
\usepackage{array}
\usepackage[caption=false,font=normalsize,labelfont=sf,textfont=sf]{subfig}
\usepackage{textcomp}
\usepackage{stfloats}
\usepackage{cite}
\usepackage{booktabs}
\usepackage{multirow}
\usepackage{threeparttable}
\usepackage{url}
\usepackage{color}

\DeclareMathAlphabet{\mathcal}{OMS}{cmsy}{m}{n}

\title{\LARGE \bf
Hierarchical Decision-Making for Autonomous Navigation: Integrating Deep Reinforcement Learning and Fuzzy Logic in Four-Wheel Independent Steering and Driving Systems
}

\author{Yizhi Wang, Degang Xu, Yongfang Xie, Shuzhong Tan, Xianan Zhou, and Peng Chen
\thanks{The authors are with the School of Automation, Central South University, Changsha 410083, China.}%
\thanks{This work was supported in part by National Natural Science Foundation of China under Grant 62473386, 61973320; in part by Key Research Project of Hunan Province, China, under Grant 2022GK2059, 2023GK2096; in part by Key Research Project of Xinjiang Province, China, under Grant 2022A02010; and in part by National Key Research and Development Program of China, under Grant 2023YFB4706900.}
\thanks{$^{\ast}$Corresponding author: Degang Xu, E-mail: {\tt\small dgxu@csu.edu.cn}}%
}

\begin{document}

\maketitle
\thispagestyle{empty}
\pagestyle{empty}

\begin{abstract}
This paper presents a hierarchical decision-making framework for autonomous navigation in four-wheel independent steering and driving (4WISD) systems. The proposed approach integrates deep reinforcement learning (DRL) for high-level navigation with fuzzy logic for low-level control to ensure both task performance and physical feasibility. The DRL agent generates global motion commands, while the fuzzy logic controller enforces kinematic constraints to prevent mechanical strain and wheel slippage. Simulation experiments demonstrate that the proposed framework outperforms traditional navigation methods, offering enhanced training efficiency and stability and mitigating erratic behaviors compared to purely DRL-based solutions. Real-world validations further confirm the framework’s ability to navigate safely and effectively in dynamic industrial settings. Overall, this work provides a scalable and reliable solution for deploying 4WISD mobile robots in complex, real-world scenarios.
\end{abstract}

\section{INTRODUCTION}
\begin{figure}[tb]
\centering
\includegraphics[width=0.9\linewidth]{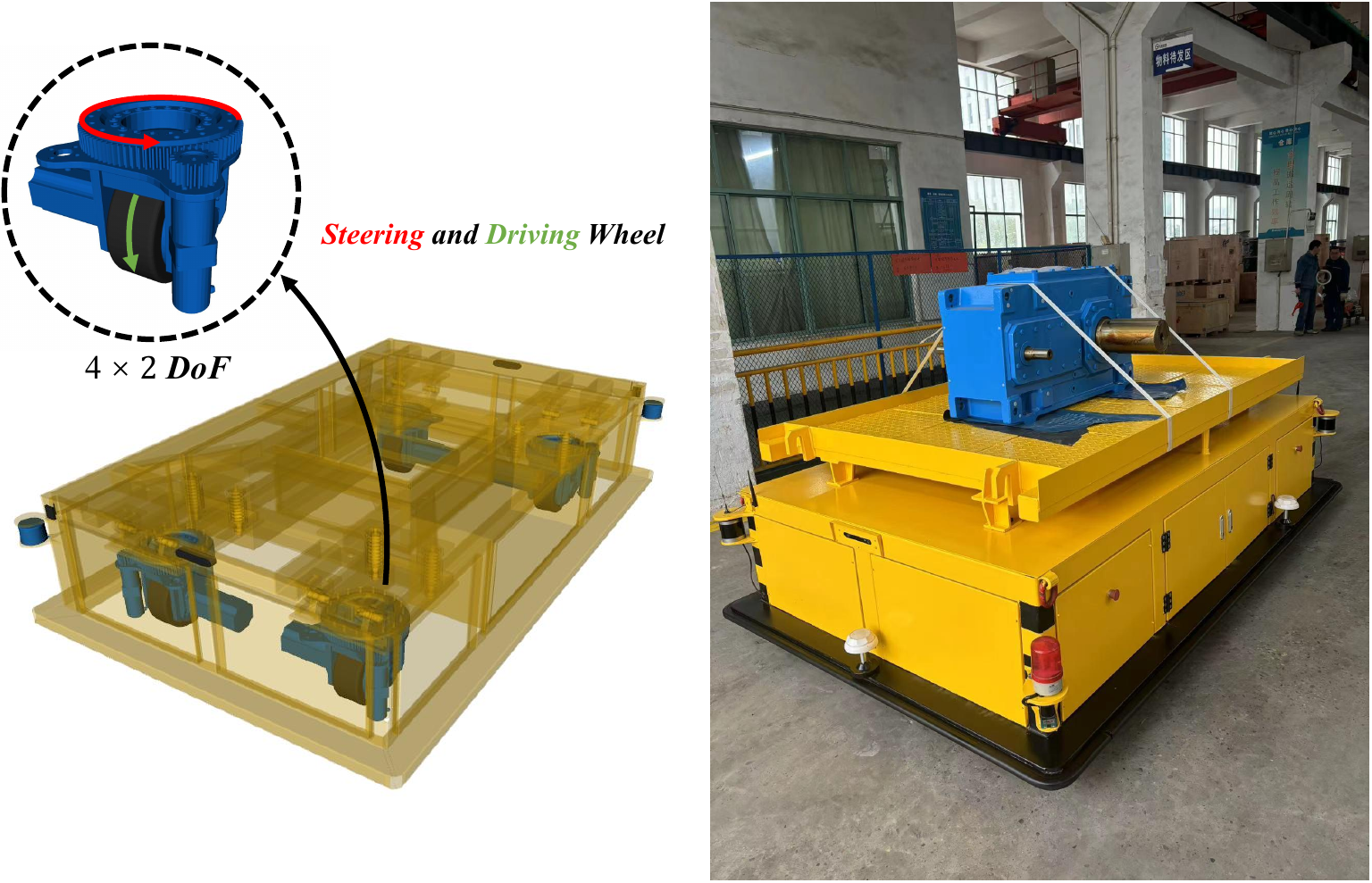}
\caption{Prototype and real-world implementation of the 4WISD AMR. Each wheel is independently steerable and drivable.}
\label{fig:intro}
\end{figure}

Autonomous mobile robots (AMRs) are increasingly deployed in industrial environments, where they perform tasks ranging from material handling to inspection and monitoring. The increasing complexity of these environments, characterized by dynamic obstacles, constrained pathways, changing layouts, and the need for real-time adaptability, demands advanced navigation and control systems \cite{xiao2022motion} that ensure safety, efficiency, and adaptability. These challenges are particularly pronounced for redundant robots, such as the four-wheel independent steering and driving (4WISD) robot, which offers unparalleled maneuverability and flexibility \cite{hang2021towards} but introduces additional control complexities due to the increased degrees of freedom and the need for precise coordination among the wheels to ensure safe and effective operation. Fig.~\ref{fig:intro} illustrates a 4WISD AMR operating in typical industrial environments.

Existing navigation solutions for mobile robots, such as Simultaneous Localization and Mapping (SLAM) and path planning algorithms  \cite{cadena2016past}, have proven effective in structured and static environments. However, their performance often degrades in dynamic and unpredictable industrial settings where conditions continuously evolve and obstacles can appear unexpectedly. Deep Reinforcement Learning (DRL) has emerged as a powerful alternative for autonomous navigation \cite{zhu2021deep}. By learning navigation policies directly from raw sensory inputs, DRL enables AMRs to adapt to new and complex environments over time. Unlike planning-based methods, DRL-based approaches allow robots to dynamically adjust to varying conditions, making them particularly well-suited for the challenges of industrial settings.

However, while DRL excels at high-level decision-making, its direct application to low-level control in redundant systems such as 4WISD robots often results in physically infeasible or suboptimal actions, such as wheel slippage or misaligned steering angles, potentially causing mechanical strain. To address these challenges, we propose a hierarchical decision-making framework for autonomous navigation in 4WISD systems. This framework integrates a high-level DRL policy for adaptive decision-making with a low-level fuzzy logic controller that enforces kinematic feasibility. The high-level policy generates global navigation commands, such as linear and angular velocities, while the low-level controller translates these commands into physically valid wheel speeds and steering angles. By combining the adaptability of learning-based methods with the reliability of model-based control, our approach ensures safe and efficient navigation in dynamic industrial environments.

To the best of the authors' knowledge, this research represents a pioneering effort to integrate deep reinforcement learning and fuzzy inference into autonomous navigation of 4WISD mobile robot, addressing the unique challenges posed by 4WISD systems, including their redundant degrees of freedom and susceptibility to kinematic misalignment. The key contributions of this work are as follows:
\begin{itemize}
\item We propose a novel hierarchical decision-making framework that combines DRL and fuzzy logic for autonomous navigation in 4WISD mobile robots.
\item We design a kinematically constrained fuzzy logic controller to minimize wheel slippage and mechanical strain, ensuring smooth and safe operation.
\item We validate the proposed framework in both simulation and real-world experiments, demonstrating its efficacy in dynamic and constrained industrial environments.
\end{itemize}

\section{RELATED WORKS}

Four-Wheel Independent Steering and Driving (4WISD) systems have transformed vehicle or mobile robot design by offering independent control over each wheel's steering and driving, significantly improving maneuverability. This technology supports complex movements like omnidirectional motion and zero-radius turns, essential for operations in dynamic, cluttered environments. Research in 4WISD for electric vehicles (EVs) has focused on enhancing stability and fault tolerance. Lam et al.\cite{lam2009omnidirectional} introduced a steer-by-wire system that minimized wheel slip, while Li et al. \cite{li2012model} developed a fault-tolerant control scheme to maintain path accuracy despite actuator failures. Further contributions by Potluri and Singh \cite{potluri2015path} improved stability with path-tracking controllers, while Kosmidis et al. \cite{kosmidis2023novel} combined neural networks with fuzzy logic to enhance robustness in 4WISD EVs.

In the field of autonomous guided vehicles (AGVs) and autonomous mobile robots (AMRs), 4WISD systems enable high-agility navigation in both indoor and outdoor environments. Setiawan et al. \cite{setiawan2016path} designed a 4WISD AGV for dynamic trajectory tracking, and Liu et al. \cite{liu2021nonlinear} employed nonlinear model predictive control for accurate path-following in difficult conditions. More recent advancements by Ding et al. \cite{ding2022trajectory} and Bae and Lee \cite{bae2023design} focused on optimizing velocity control and adaptive steering systems, enhancing maneuverability in confined spaces. These systems are particularly valuable for AGVs used in applications like heavy-duty transport, where both precision and flexibility are essential.

Traditional navigation methods, such as SLAM, path planning, and obstacle avoidance techniques, have been foundational in autonomous systems but are often limited in dynamic, unstructured environments. Advances have been made for omnidirectional and industrial robots, including Sprunk et al. \cite{sprunk2017accurate}, who developed trajectory generation and velocity planning optimized for industrial settings, and Shin et al. \cite{shin2021model}, who introduced model predictive path planning for rough terrain navigation. More recent work by Ma et al. \cite{ma2022bi} and Yilmaz et al. \cite{yilmaz2022precise} integrated risk assessments and high-precision localization to improve navigation in unpredictable environments. Despite these efforts, traditional methods still face challenges in adapting to evolving layouts and handling unexpected obstacles.

Deep Reinforcement Learning has gained attention for autonomous navigation, enabling robots to adapt through interaction with their environment. Early works \cite{zhu2017target, tai2017virtual, fan2018crowdmove} demonstrated DRL's potential in dynamic environments, with approaches like target-driven models and mapless navigation using sparse LiDAR data. Recent developments have focused on enhancing DRL efficiency and generalization. Jang et al.\cite{jang2021hindsight} introduced Hindsight Intermediate Targets (HIT) to enhance learning efficiency, while Zhu et al.\cite{zhu2022hierarchical} proposed a hierarchical DRL framework for safer navigation. Reward shaping techniques are employed by works \cite{miranda2023generalization, guo2023optimal, wang2024heuristic} to improve navigation performance and generalization. Integrating DRL with 4WISD systems offers significant potential, as the versatile movement capabilities of 4WISD enhance the ability of DRL-based frameworks to navigate complex and dynamic environments.

\section{Problem Statement}

The problem of autonomous navigation for a 4WISD AMR is decomposed into two hierarchical kinematic levels: macro kinematics for high-level navigation in the task space and micro kinematics for low-level redundancy resolution in control space.

\begin{figure*}[t]
    \centering
    \subfloat[]{\includegraphics[width=0.75\linewidth]{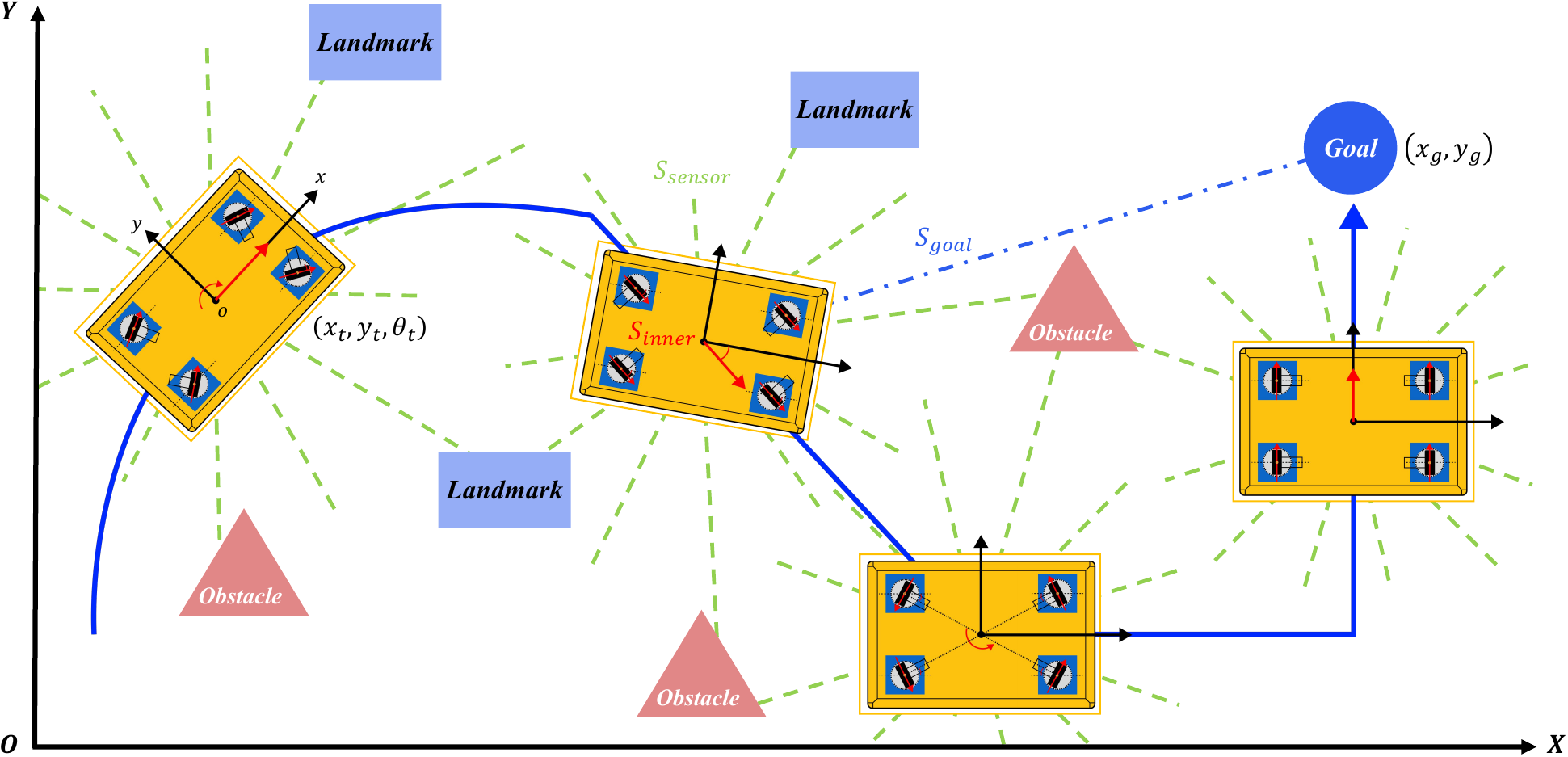}\label{fig:macro_kinematics}}\hfill
    \subfloat[]{\includegraphics[width=0.25\linewidth]{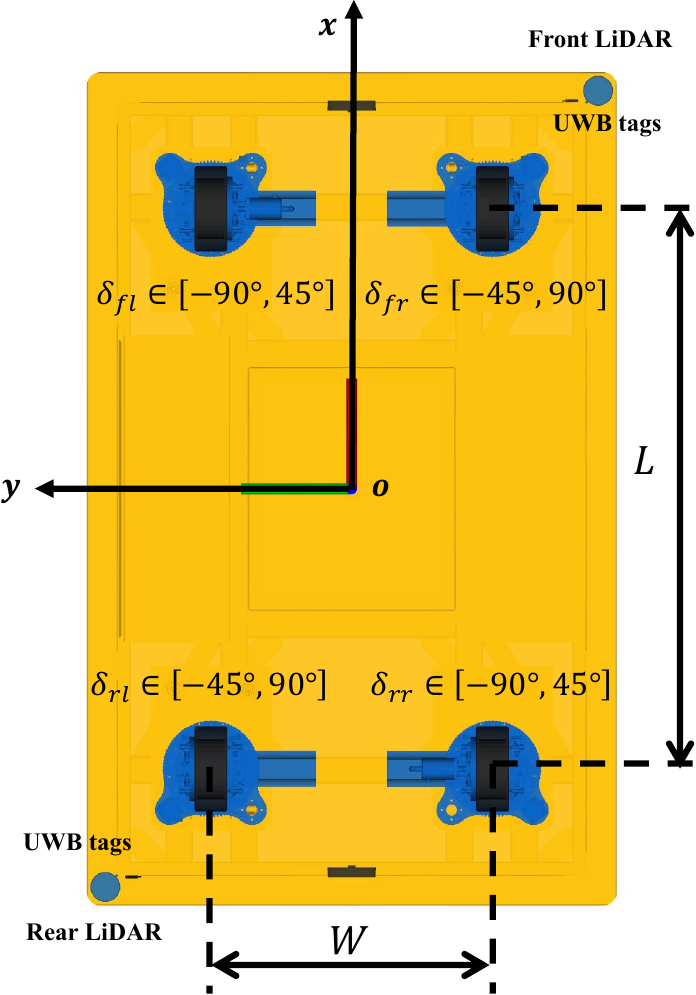}\label{fig:micro_kinematics}}
    \caption{(a) The AMR navigates from its initial pose $(x_{t}, y_{t}, \theta_{t})$ to a target $(x_{g}, y_{g})$ in the global coordinate system $XOY$. (b) The robot's configuration in the local coordinate system \(xoy\), including sensor placement, geometric dimensions (wheelbase $L$, steering track $W$), steering angles $\delta_{fl}, \delta_{fr}, \delta_{rl}, \delta_{rr}$ and wheel velocities $v_{fl}, v_{fr}, v_{rl}, v_{rr}$.}
\end{figure*}

\subsection{Macro Kinematics for Navigation}
The macro kinematics governs the robot's high-level movement in a global coordinate system $XOY$, where its pose at time $t$ is defined as $(x_t, y_t, \theta_t)$, (see Fig.~\ref{fig:macro_kinematics}). The robot aims to navigate from navigate from $(x_t, y_t)$ to a target $(x_g, y_g)$ while avoiding obstacles. Its motion is modeled by the kinematic equation:
\begin{equation}
\begin{bmatrix}\dot{x_t}\\
               \dot{y_t}\\
               \dot{\theta_t}\end{bmatrix}
               =
               \begin{bmatrix}\cos{\theta_t} &-\sin{\theta_t} &0\\
                              \sin{\theta_t} &\cos{\theta_t} &0\\
                              0 &0 &1\end{bmatrix}
               \begin{bmatrix}v_{x}\\
               v_{y}\\
               \omega_{z}\end{bmatrix},
\end{equation}
where $v_x$ and $v_y$ are the linear velocities, and $\omega_z$ represents the angular velocity in the robot’s local frame $xoy$. The navigation task is formulated as a Markov Decision Process (MDP), where the state space $\mathcal{S}$, action space $\mathcal{A}$, transition function $P$, reward function $R$, and discount factor $\gamma$ are defined. At each time step $t$, the robot observes a state $s_{t} \in \mathcal{S}$ and selects an action $a_{t} \in \mathcal{A}$ based on its policy $\pi: \mathcal{S} \rightarrow \mathcal{A}$, aiming to maximize the cumulative discounted rewards:
\begin{equation}
    \pi^\ast=\mathop{\arg\max}\limits_{\pi} \mathop{\mathbb{E}}_{(s_{t}, a_{t})\sim\rho_{\pi}}\left[\sum_{\tau=0}^{\infty }\gamma^{\tau}r_{t+\tau}\right],
\end{equation}
where $\gamma$ is the discount factor and $\rho_{\pi}$ is the distribution over state-action pairs induced by the policy $\pi$. 

\subsection{Micro Kinematics for Control}

The micro kinematics resolves the robot’s redundant actuation in the local coordinate system $xoy$, where its motion is governed by four independently steerable and driven wheels (Fig.~\ref{fig:micro_kinematics}). The task-space velocities are related to the wheel velocities $v_i$ and steering angle $\delta_i$ ($i \in \{fl, fr, rl, rr\}$) through the kinematic constraints: 
\begin{equation}
\begin{bmatrix}
v_x\\
v_y\\
\omega_z
\end{bmatrix}
= \\
\mathbf{J}(\delta_{fl}, \delta_{fr}, \delta_{rl}, \delta_{rr})
\begin{bmatrix}
v_{fl}\\
v_{fr}\\
v_{rl}\\
v_{rr}
\end{bmatrix}, \\
\end{equation}
where the Jacobian matrix $\mathbf{J} \in \mathbb{R}^{3 \times 4}$ depends on the wheelbase $L$, steering track $W$, and the steering angles $\delta_{fl}, \delta_{fr}, \delta_{rl}, \delta_{rr}$. Given the redundancy in control inputs (more control variables than degrees of freedom), the system is underdetermined, making it impossible to directly invert the Jacobian matrix. Therefore, redundancy resolution is necessary to determine feasible wheel velocities that satisfy the robot’s kinematics.

\section{Hierarchical Decision-Making Framework}

\begin{figure*}[tb]
\centering
\includegraphics[width=\linewidth]{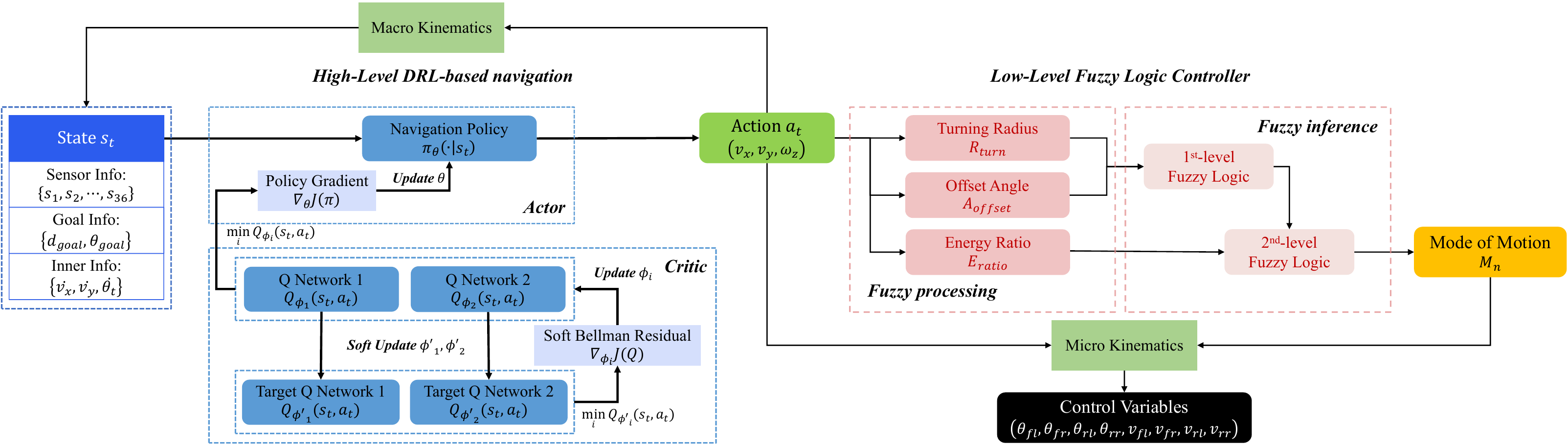}
\caption{Overview of the hierarchical decision-making framework. The high-level DRL-based navigation module learns an optimal policy by interacting with the environment. The actor generates actions based on the current state, while the critic evaluates the Q-value. The optimization process leverages entropy-regularized policy gradients to update the policy and the soft Bellman equation to optimize the Q-function. The low-level fuzzy logic controller translates the velocity commands from the DRL policy, ensuring kinematic feasibility by selecting appropriate motion modes. It then computes the final wheel velocities and steering angles required for smooth and safe navigation of the 4WISD system.}
\label{fig:overview}
\end{figure*}

In order to effectively address the challenges posed by the two distinct kinematic levels, the proposed hierarchical decision-making framework integrates a high-level DRL-based navigation policy with a low-level fuzzy logic controller, as shown in Fig.~\ref{fig:overview}. This hierarchical structure enables the 4WISD system to navigate effectively while maintaining stability and safety.

\subsection{High-Level DRL-based navigation}

The high-level navigation utilizes a Deep Reinforcement Learning (DRL) approach, with the Soft Actor-Critic (SAC) algorithm \cite{haarnoja2018soft} selected for its capability to handle continuous action spaces and promote policy robustness through entropy regularization. SAC ensures stability and efficiency during learning, making it particularly suitable for the robot's navigation task in dynamic and uncertain environments. 

\subsubsection{State Representation and Action Space}

The state $s_t$ provides a comprehensive representation of the robot's operational context, integrating environmental observations, goal information, and internal dynamics:
\begin{equation}
s_t = \{s_{\text{sensor}}, s_{\text{goal}}, s_{\text{inner}}\}.
\end{equation}
The sensor state, $s_{\text{sensor}}$, is derived from dual LiDARs capturing a 2D spatial scan, represented as $s_{\text{sensor}} = \{s_1, s_2, \dots, s_{36}\}$. Each element $s_{i}$ corresponds to a nonlinear normalized sample point in the LiDAR scan for obstacle detection. The goal state $s_{\text{goal}} = (d_{\text{goal}}, \theta_{\text{goal}} )$ includes the robot’s Euclidean distance to the goal $d_{\text{goal}}$ and its angular orientation $\theta_{\text{goal}}$ relative to its heading:
\begin{equation}
\begin{aligned}
d_{\text{goal}}      &= \sqrt{(x_g - x_t)^2 + (y_g - y_t)^2}, \\
\theta_{\text{goal}} &= \arctan2(y_g - y_t, x_g - x_t) - \theta_t.
\end{aligned}
\end{equation}
The robot’s internal dynamics are captured through IMU data, which measures the linear accelerations along the x- and y-axes and angular velocity around the z-axis. The inner state, $s_{\text{inner}} = \{\dot{v_x}, \dot{v_y}, \dot{\theta_t}\}$, provides dynamic feedback on the robot's motion.

The action space is defined by three continuous commands representing the robot’s linear velocities $v_x, v_y$ and angular velocity $\omega_z$:
\begin{equation}
a_t=(v_x, v_y, \omega_z).
\end{equation}
These actions are based on macro kinematics, ensuring that physical movements are consistent with the robot's capabilities.

\subsubsection{Behavior-based Reward Function}

The reward function promotes behavior aligned with the operational objectives: minimizing the distance to the goal, ensuring safe navigation, and maintaining stability. The total reward at time step $t$ is the sum of three components:
\begin{equation}
r_t = R_{\text{progress}} + R_{\text{safety}} + R_{\text{stability}}.
\end{equation}

The Progress Reward $R_{\text{progress}}$ encourages the robot to reduce its distance to the goal and is computed as:
\begin{equation}
R_{\text{progress}} = \lambda_{\text{progress}} \cdot \Delta d_{\text{goal}},
\end{equation}
where $\Delta d_{\text{goal}}$ is the change in Euclidean distance between consecutive time steps, and $\lambda_{\text{progress}}$ is a scaling factor.

To ensure that the robot navigates safely without colliding with obstacles, the Safety Reward $R_{\text{safety}}$ penalizes proximity to obstacles when the robot is too close. It is defined as:
\begin{equation}
R_{\text{safety}} = 
\begin{cases} 
    -\lambda_{\text{safety}} \cdot (r_{\text{safe}} - r_{\text{obs}}), & \text{if } r_{\text{obs}} < r_{\text{safe}}, \\
    0, & \text{otherwise},
\end{cases}
\end{equation}
where $r_{\text{obs}}$ is the distance to the nearest obstacle, $r_{\text{safe}}$ is the minimum safety distance, and $\lambda_{\text{safety}}$ is a penalty factor.
    
The Stability Reward $R_{\text{stability}}$ encourages smooth and stable motion by penalizing large fluctuations in the measured accelerations and angular velocity. The reward component is calculated as:
\begin{equation}
R_{\text{stability}} = -\lambda_{\text{stability}} \cdot (\dot{v_x}^2 + \dot{v_y}^2 + |\dot{\theta_t}|),
\end{equation}
where $\lambda_{\text{stability}}$ is a weighting factor.

\subsubsection{Neural Network Architecture}

The SAC algorithm is implemented using an actor-critic architecture, where both the actor and critic networks are fully connected neural networks. This structure ensures computational efficiency while providing sufficient representational capability for the navigation task.

\textbf{Actor network}, $\pi_{\theta}(\cdot \vert s_t)$, learns a stochastic policy mapping the state $s_t$ to a distribution over actions, parameterized by parameter $\theta$. This network comprises an input layer corresponding to the state dimensions, followed by two hidden layers with 512 neurons and ReLU activations. The output layer has dimensions corresponding to the action space, with separate outputs for the mean ($\mu$) and standard deviation ($\sigma$) of the action distribution. The mean is constrained to $[-1, 1]$ using a \textit{tanh} activation, while the standard deviation is modeled as a positive value using a \textit{softplus} activation.

The actor's output is scaled to match the robot’s kinematic limits:
\begin{equation}
\begin{aligned}
v_x &= v_x^{\text{min}} 
      + \tfrac{\mu_{v_x} + 1}{2}(v_x^{\text{max}} - v_x^{\text{min}}), \\
v_y &= v_y^{\text{min}} 
      + \tfrac{\mu_{v_y} + 1}{2}(v_y^{\text{max}} - v_y^{\text{min}}), \\
\omega_z &= \omega_z^{\text{min}} 
      + \tfrac{\mu_{\omega_z} + 1}{2}(\omega_z^{\text{max}} - \omega_z^{\text{min}}).
\end{aligned}
\end{equation}

\textbf{Critic networks}, $Q_{\phi}(s_t, a_t)$, estimate the soft Q-values for state-action pairs, parameterized by parameter $\phi$. The critic networks use a similar architecture to the actor network, with an additional input dimension for actions and a scalar output for the Q-value. This scalar represents the expected return for the given state-action pair under the current policy. 

The simplicity of the actor-critic networks is purposeful—complex architectures are not critical for our application as long as the network reliably maps the non-linear relationships inherent in the problem.

\subsubsection{Training and Optimization}

The training process involves iteratively optimizing the policy and Q-functions. The policy objective maximizes the expected return, augmented with an entropy term $H(\pi(\cdot \vert s_{t}))$ to encourage exploration:
\begin{equation}
  J(\pi) = \mathop{\mathbb{E}}_{(s_{t}, a_{t}) \sim \rho_{\pi}} \left[\sum_{t=0}^{T} \gamma^{t} \left(r_{t} + \alpha H(\pi(\cdot \vert s_{t}))\right)\right],
\end{equation}
where $H(\pi(\cdot \vert s_{t})) = \mathbb{E}_{a_{t} \sim \pi}[-\log \pi(a_{t} \vert s_{t})]$ represents the entropy of the policy, and $\alpha$ is the temperature parameter balancing exploration and exploitation.

The Q-functions $Q_{\phi_1}$ and $Q_{\phi_2}$ are optimized using the soft Bellman residual\cite{haarnoja2017reinforcement}:
\begin{equation}
J(Q) = \mathop{\mathbb{E}}_{(s_t, a_t, r_t, s_{t+1})} \left[ \left(Q_{\phi_i}(s_t, a_t) - y_t\right)^2 \right],
\end{equation}
where the target value $y_t$ is computed using the minimum of the two corresponding target Q-functions $Q_{\phi_1'}$ and $Q_{\phi_2'}$:
\begin{equation}
y_t = r_t + \gamma \mathop{\mathbb{E}}_{a_{t+1} \sim \pi} \left[ \min_{i=1,2} Q_{\phi_i'}(s_{t+1}, a_{t+1}) - \alpha \log \pi_{\theta}(a_{t+1} \vert s_{t+1}) \right].
\end{equation}

The policy is updated using the entropy-regularized gradient:
\begin{equation}
\nabla_{\theta} J(\pi) = \mathop{\mathbb{E}}_{a_t \sim \pi} \left[ \alpha \nabla_{\theta} \log \pi_{\theta}(a_t \vert s_t) - \nabla_{\theta} Q_{\phi}(s_t, a_t) \right].
\end{equation}

The use of two Q-functions reduces overestimation bias, while target networks ensure training stability via soft updates:
\begin{equation}
\phi_i' \leftarrow \tau \phi_i + (1 - \tau) \phi_i',
\end{equation}
where $\tau$ is a smoothing coefficient.

\subsection{Low-Level Fuzzy Logic Controller}

The low-level fuzzy logic controller translates high-level navigation commands into feasible control signals for the 4WISD system. This controller addresses the underdetermined inverse kinematics problem, where the number of control inputs exceeds the available kinematic equations. It ensures both kinematic alignment and safe operation by managing wheel velocities and steering angles to minimize mechanical strain, wheel slippage, and internal forces. 

The controller transforms the velocity command $(v_{x}\ v_{y}\ \omega_{z})$ into three fuzzy linguistic variables: turning radius $R_{\text{turn}}$, velocity offset angle $A_{\text{offset}}$, and energy ratio $E_{\text{ratio}}$. These variables are fed into a two-level fuzzy inference system that determines the appropriate mode of motion $M$. Based on $M$, the controller applies corresponding micro-kinematic equations to compute the individual wheel steering angles and velocities $(\delta_{fl}\ \delta_{fr}\ \delta_{rl}\ \delta_{rr}\ v_{fl}\ v_{fr}\ v_{rl}\ v_{rr})$. 

\subsubsection{Mode-based constraints}

\begin{figure*}[tb]
\centering
\includegraphics[width=\linewidth]{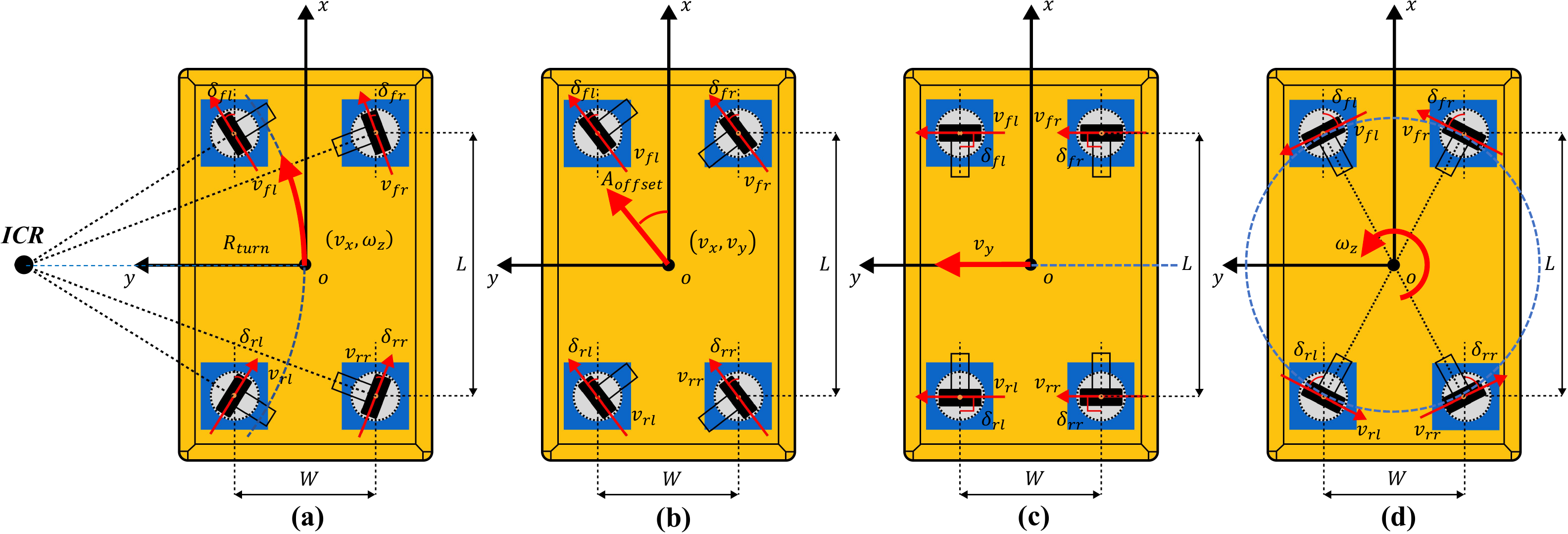}
\caption{Representation of the four motion modes used in the 4WISD system: (a) Steering, (b) Oblique, (c) Lateral, and (d) Rotation. Each mode defines specific constraints on wheel velocities and steering angles, enabling efficient robot control in diverse operational scenarios.}
\label{fig:mode}
\end{figure*}

The robot motion is governed by four distinct modes, each imposing specific constraints on wheel velocities and steering angles (see Fig.~\ref{fig:mode}).

\textbf{Steering Mode (SM)}: Designed for conventional turning, this mode assumes that the instantaneous center of rotation (ICR) lies along the robot’s transverse symmetry axis. The minimum turning radius is determined as $R_{\text{min}} = (W \tan(\pi/2 - |\delta|_{\text{min}}) + L)/2$. Steering angles are computed to align the wheels with the curvature of motion, and wheel velocities are adjusted accordingly to maintain coordinated movement.

\textbf{Oblique Mode (OM)}: In this mode, all wheels align at a common steering angle, enabling diagonal movement without changing the robot’s heading. The velocity direction is offset by an angle $\alpha = \arctan(v_y / v_x)$, where $|\alpha| \leq |\delta_{\text{min}}|$, ensuring kinematic feasibility. 

\textbf{Lateral Mode (LM)}: By constraining all wheels to a $90^\circ$ steering angle, the robot achieves pure sideways movement. This mode is especially useful for tasks like sideways docking and fine adjustments in confined spaces.

\textbf{Rotation Mode (RM)}: This mode enables in-place rotation about the robot’s geometric center by symmetrically adjusting steering angles. The effective spin radius is given by $R_{\text{spin}} = \sqrt{(L/2)^2 + (W/2)^2}$.

These modes simplify the complexity of redundant control by constraining the wheel configurations, ensuring that movements remain kinematically valid. However, dynamically selecting the optimal mode introduces new challenges, which are addressed by the fuzzy inference system.

\subsubsection{Input and Output Variables}

The fuzzy inference system utilizes three key input linguistic variables, derived from physical and mathematical calculations.

\textbf{Turning Radius ($R_{\text{turn}}$)}: Represents the curvature of the robot's path, calculated as:
    \begin{equation}
    R_{\text{turn}} = \frac{v_x}{\omega_z}.
    \end{equation}
    Fuzzy subsets are Radius-Unrealized (RU), Radius-Realized (RR), and Radius-Infinity (RI).
    
\textbf{Velocity Offset Angle ($A_{\text{offset}}$)}: Measures the deviation of the linear velocity vector from the forward direction:
    \begin{equation}
    A_{\text{offset}} = \arctan\left(\frac{v_y}{v_x}\right).
    \end{equation}
    Fuzzy subsets include Angle-Zero (AZ), Angle-Reachable (AR), Angle-Unreachable (AU), and Angle-Limited (AL).
    
\textbf{Energy Ratio ($E_{\text{ratio}}$)}: Represents the balance between lateral kinetic energy $E_y$ and rotational kinetic energy $E_z$, analogous to the turning radius:
    \begin{equation}
    E_{\text{ratio}} = \frac{v_y}{\omega_z}.
    \end{equation}
    where $E_y = (M v_y^2)/2$ and $E_z = (I \omega_z^2)/2$ are the kinetic energy contributions, and $I = M(L^2 + W^2)/12$ is the moment of inertia. Fuzzy subsets include Rotation-Leading (RL) and Translation-Leading (TL).

\begin{figure}[tb]
\centering
\includegraphics[width=\linewidth]{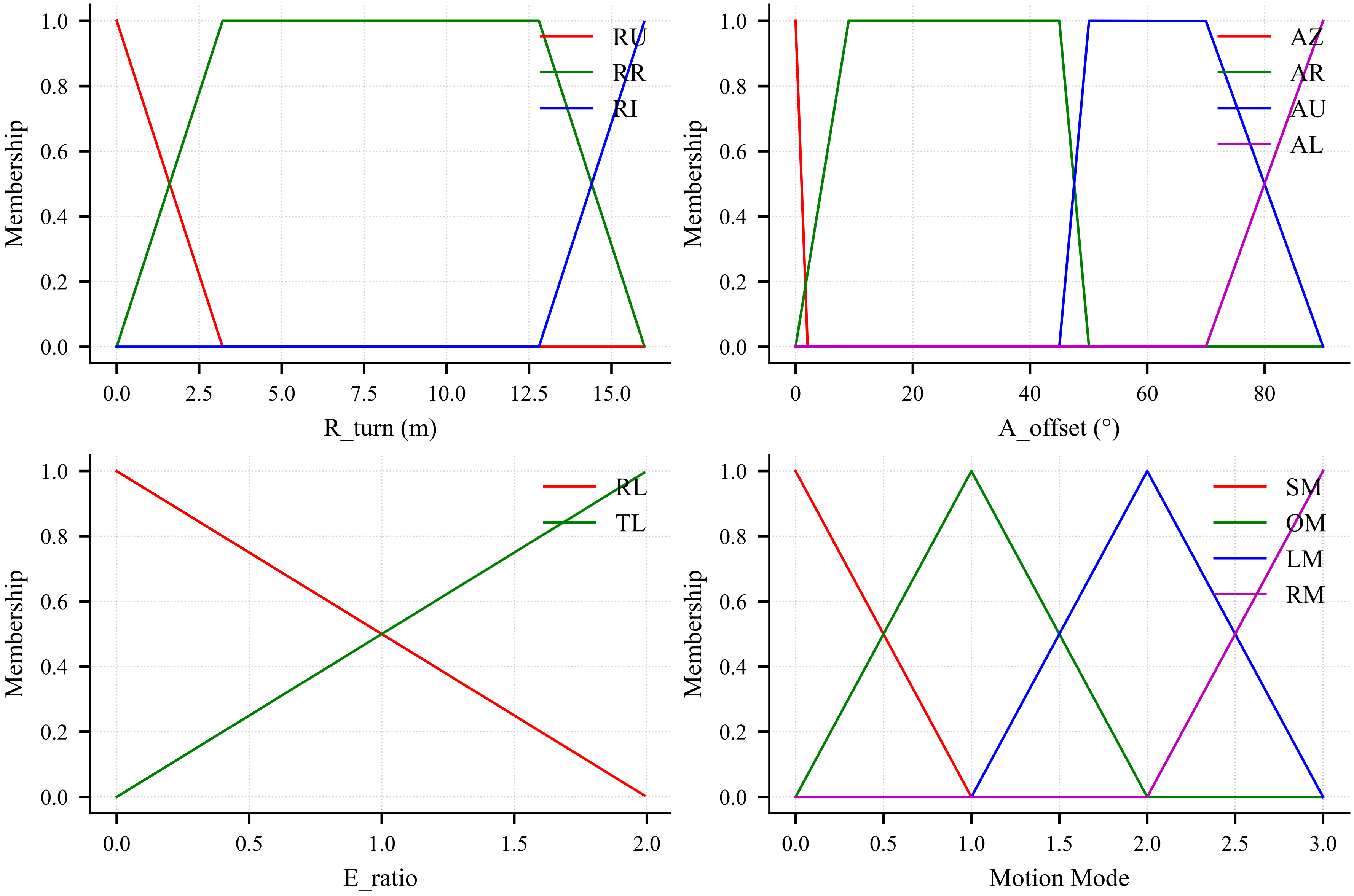}
\caption{Membership functions for turning radius, velocity offset angle, energy ratio, and mode of motion.}
\label{fig:msf}
\end{figure}

The output linguistic variable is the \textbf{Mode of Motion ($M$)}, which corresponds to one of four modes: SM, OM, LM, or RM. The membership functions for these input and output variables are defined using triangular and trapezoidal shapes, as depicted in Fig.~\ref{fig:msf}. The triangular functions simplify computation, while trapezoidal functions offer wider stable regions of membership, reducing sensitivity to minor input variations.

\subsubsection{Fuzzy inference and Defuzzification}

The fuzzy inference system relies on a set of expert rules to determine the robot's motion mode. These fuzzy rules are organized in two levels, as summarized in Table~\ref{tab:fuzzy_rule}. The first-level establishes broad motion tendencies based on $R_{\text{turn}}$ and $A_{\text{offset}}$, while the second-level refines the decision process by factoring in $E_{\text{ratio}}$ to determine the most appropriate mode.

\begin{table}[h]
\centering
\caption{The table of Fuzzy logic rule.}
\label{tab:fuzzy_rule}
\begin{tabular}{l|c|c|c|c}
\hline
\textbf{$1^{st}$-level} & \textbf{AZ} & \textbf{AR} & \textbf{AU} & \textbf{AL}\\
\hline
\textbf{RI} & SM & OM  & OM & LM\\
\textbf{RU} & RM & SM  & SM & RM\\
\textbf{RR} & SM & SM  & SM & SM\\
\hline
\textbf{$2^{nd}$-level} & SM & OM & LM & RM\\
\hline
\textbf{RL} & SM  & OM & SM & RM\\
\textbf{TL} & OM  & OM & LM & LM\\
\hline
\end{tabular}
\end{table}

The fuzzy output $M^*$ is determined using Mamdani reasoning method:
\begin{equation}
\begin{aligned}
    M^* &= \left( \left( (R^* \wedge A^*) \vee F_1 \right) \wedge E^* \right) \vee F_2,
\end{aligned}
\end{equation}
where $R^*$, $A^*$ and $E^*$ denote the fuzzy membership degrees corresponding to the turning radius, velocity offset angle, and energy ratio, respectively. The symbol $\wedge$ represents the minimum operator (fuzzy intersection) and $\vee$ denotes the maximum operator (fuzzy union). The inference steps $F_1$ and $F_2$ are computed as:
\begin{equation}
\begin{aligned}
    F_1 &= \bigvee\limits_{i=1, j=1}^{i=3, j=4} \left\{\left( \mu_R(R_i) \wedge \mu_A(A_j) \right) \wedge \mu_{M^1}(M_{ij}^1)\right\} \\
    F_2 &= \bigvee\limits_{n=1, m=1}^{n=4, m=2} \left\{\left( \mu_{M^1}(M_n^1) \wedge \mu_E(E_m) \right) \wedge \mu_{M^2}(M_{nm}^2)\right\}
\end{aligned}
\end{equation}

The defuzzification process, based on the Mean of Maximum (MOM) method, determines the mode of motion by selecting the value of $M_n$ that maximizes the membership function $\mu_M(M_n)$:
\begin{equation}
    M_n = \text{argmax}_{n=1,2,3,4} \mu_M(M_n)
\end{equation}
where $\text{argmax}$ denotes the argument that maximizes the membership function $\mu_M(M_n)$.

\section{Experiments and Results}

\subsection{Simulation Experiments}

Simulation experiments were conducted to evaluate the navigation performance and generalization capability of the proposed hierarchical decision-making framework. The framework was compared against the Dynamic Window Approach (DWA) \cite{fox1997dynamic} and Timed Elastic Band (TEB) \cite{rosmann2015timed}, two widely used planning-based navigation methods, both of which also incorporate our fuzzy logic controller.

To ensure a comprehensively comparison, all methods were tested in three distinct scenarios with different navigation tasks. The evaluation focused on four key performance metrics: pose precision (PP), which measures the final positional error relative to the goal; average speed (AS), which reflects the efficiency of movement; path efficiency (PE), defined as the ratio of the planned optimal path length to the actual traversed path, indicating trajectory efficiency; and success rate (SR), which represents the percentage of successful goal-reaching attempts without failure.

\renewcommand{\thetable}{\Roman{table}}

\begin{table}[htbp]
\centering
\caption{Simulation Results Across Three Scenarios.}
\label{tab:simulation_results}
\begin{tabular}{c|c|c|c|c|c}
\hline
\textbf{Env} & \textbf{Method} & \textbf{PP} (m) & \textbf{AS (m/s)} & \textbf{PE (\%)} & \textbf{SR (\%)} \\ 
\hline
\multirow{3}{*}{S1} 
    & DWA       & 0.184           & 0.358           & \textbf{98.7}  & 90.0 (27/30) \\ 
    & TEB       & \textbf{0.097}  & 0.437           & 92.4           & 73.3 (22/30) \\ 
    & Proposed  & 0.199           & \textbf{0.506}  & 74.8           & \textbf{96.7(29/30)} \\ 
\hline
\multirow{3}{*}{S2} 
    & DWA       & 0.185           & 0.384           & \textbf{99.5}  & 84.8 (28/33) \\ 
    & TEB       & \textbf{0.068}  & 0.478           & 97.4           & 87.9 (29/33) \\ 
    & Proposed  & 0.155           & \textbf{0.616}  & 90.6           & \textbf{100.0 (33/33)} \\ 
\hline
\multirow{3}{*}{S3} 
    & DWA       & 0.187           & 0.424           & \textbf{102.2} & 66.7 (18/27) \\ 
    & TEB       & \textbf{0.116}  & 0.489           & 99.0          & 85.2 (23/27) \\ 
    & Proposed  & 0.161           & \textbf{0.558}  & 86.1           & \textbf{92.6 (25/27)} \\ 
\hline
\end{tabular}
\end{table}

Table~\ref{tab:simulation_results} summarizes the results across three test scenarios (S1, S2, S3). The proposed framework achieves slightly larger but competitive pose precision compared to DWA, while consistently reaching the goal within the 0.2 m training tolerance in all cases. It achieves the highest AS across all scenarios, demonstrating improved movement efficiency over DWA and TEB, which are constrained by their reactive and local trajectory adjustments.

In terms of PE, DWA maintains the most efficient paths due to its velocity-space-based optimization, while TEB follows closely but occasionally compromises efficiency for stricter orientation tracking. The proposed framework exhibits the lowest PE, which can be attributed to its decision-making based navigation that prioritizes safe and adaptable movement over strictly optimal shortest paths.

Notably, the proposed framework consistently achieves the highest SR, reaching 100\% in S2 and outperforming DWA and TEB in all cases. This highlights its robustness and generalization ability, effectively reducing failure cases caused by stuck conditions or obstacle collisions. Unlike DWA and TEB, which struggle in certain environments due to local minima or excessive path corrections, the hierarchical framework dynamically adjusts its control strategy, enabling safer and more reliable navigation.

Overall, these results confirm that the hierarchical framework balances navigation success, efficiency, and adaptability, making it a robust alternative to conventional planning-based approaches. While it does not strictly optimize path length, its emphasis on safe and stable execution ensures improved real-world feasibility across diverse scenarios.

\subsection{Ablation Study}

The ablation study was conducted to assess the impact of the hierarchical decision-making framework on learning efficiency and navigation performance. Specifically, we compared the proposed hierarchical framework to a pure DRL approach in which the SAC model directly outputs wheel velocities and steering angles without explicit kinematic constraints. Both methods were trained using the same state representation and reward structure, with hyperparameters listed in Table~\ref{tab:hyperparameters}.

\begin{table}[htbp]
\centering
\caption{Hyperparameter Settings for Training}
\label{tab:hyperparameters}
\begin{tabular}{lc|lc}
\hline
\multicolumn{2}{c|}{\textbf{Training Parameter}}         & \multicolumn{2}{c}{\textbf{Robotics Parameter}} \\
\hline
\textbf{Parameter}            & \textbf{Value}           & \textbf{Parameter}        & \textbf{Value}      \\
\hline
Training episodes         & 2000                     & \(\lambda_{\text{progress}}\)  & 0.5                 \\
Buffer size            & \(5 \times 10^5\)          & \(\lambda_{\text{safety}}\)    & 0.3                 \\
Batch size                    & 128                      & \(\lambda_{\text{stability}}\) & 0.5                 \\
Episode length                & 100 steps                & \(v_{x}^{\text{max}}\)         & 0.75\,m/s          \\
Learning rate (Actor)         & \(3 \times 10^{-4}\)       & \(v_{y}^{\text{max}}\)         & 0.35\,m/s          \\
Learning rate (Critic)        & \(3 \times 10^{-4}\)       & \(\omega_{z}^{\text{max}}\)    & 0.32\,rad/s        \\
Discount factor (\(\gamma\))  & 0.99                     & Wheelbase (\(L\))            & 2.03\,m             \\
Soft update (\(\tau\)) & 0.005                    & Steer track (\(W\))          & 1.02\,m             \\
\hline
\end{tabular}
\end{table}

\begin{figure}[ht]
    \centering
    \includegraphics[width=\linewidth]{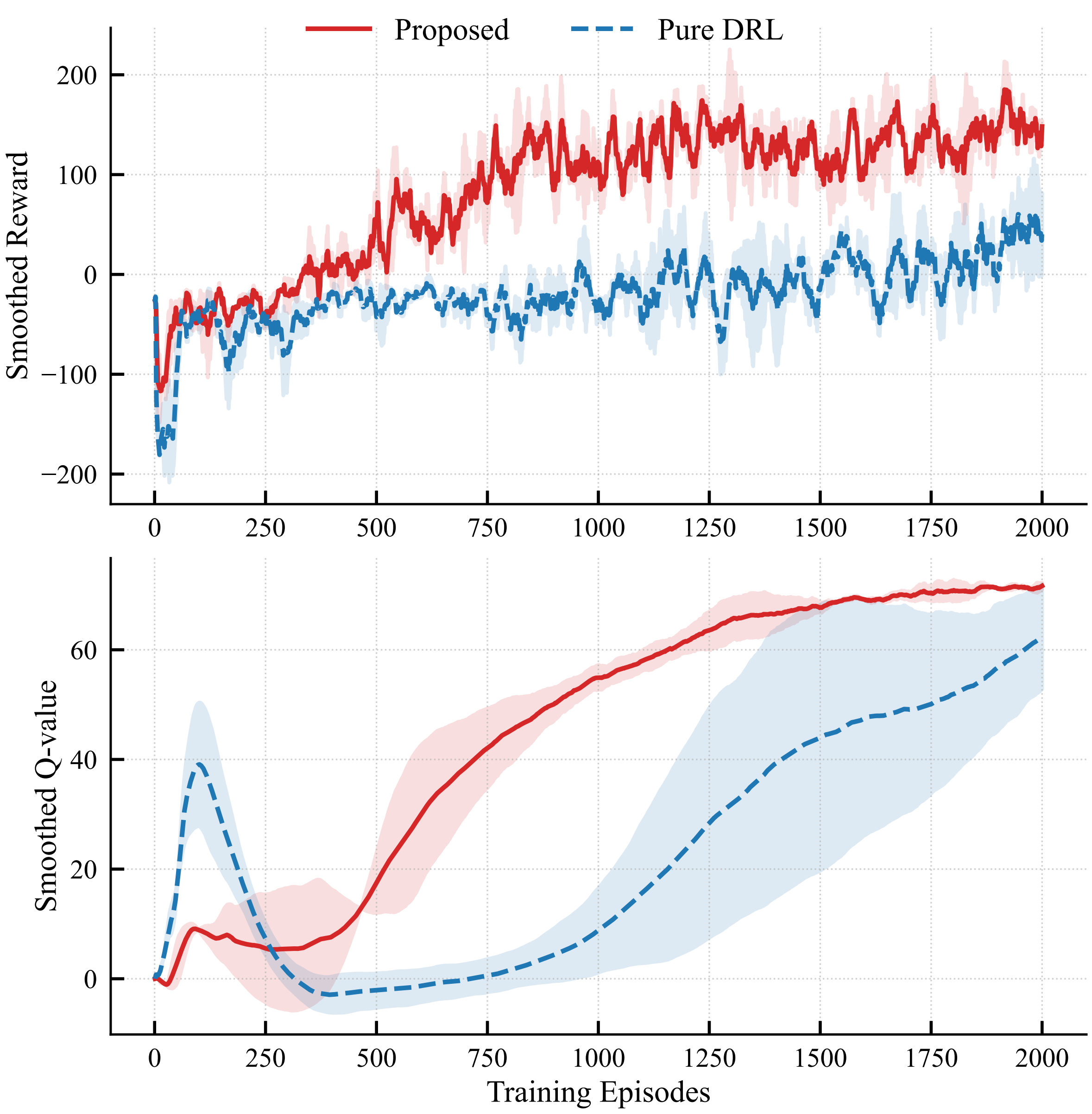}
    \caption{Training curves comparing the pure DRL approach and the proposed hierarchical framework. The hierarchical model demonstrates faster convergence and improved training stability, whereas the pure DRL approach exhibits Q-value overestimation and fluctuations in early training.}
    \label{fig:training_curves}
\end{figure}

Fig.~\ref{fig:training_curves} presents the training curves for both methods. The pure DRL approach exhibits Q-value overestimation during early training, which leads to unstable and erratic behaviors. These erratic motion patterns, characterized by sharp turns, wheel dragging and excessive friction, result in inefficient exploration of the environment. The frequent occurrence of suboptimal actions restricts the diversity of experiences, thus hindering effective policy learning.

In contrast, integrating the fuzzy logic controller in the hierarchical framework enforces kinematic constraints, resulting in smoother and more controlled actions. This constrained action space improves the quality of exploration by preventing extreme, physically unfeasible maneuvers. As a consequence, the agent gathers more informative experiences during training, leading to faster convergence and consistently higher average episode rewards.

Although the pure DRL-based policy eventually allows the robot to reach the target, its erratic motion patterns cause mechanical stress and compromise stability. Conversely, the hierarchical framework achieves better navigation performance under identical training conditions and mitigates issues related to excessive mechanical strain.

\subsection{Real-world Experiments}
\begin{figure}[ht]
\centering
\includegraphics[width=\linewidth]{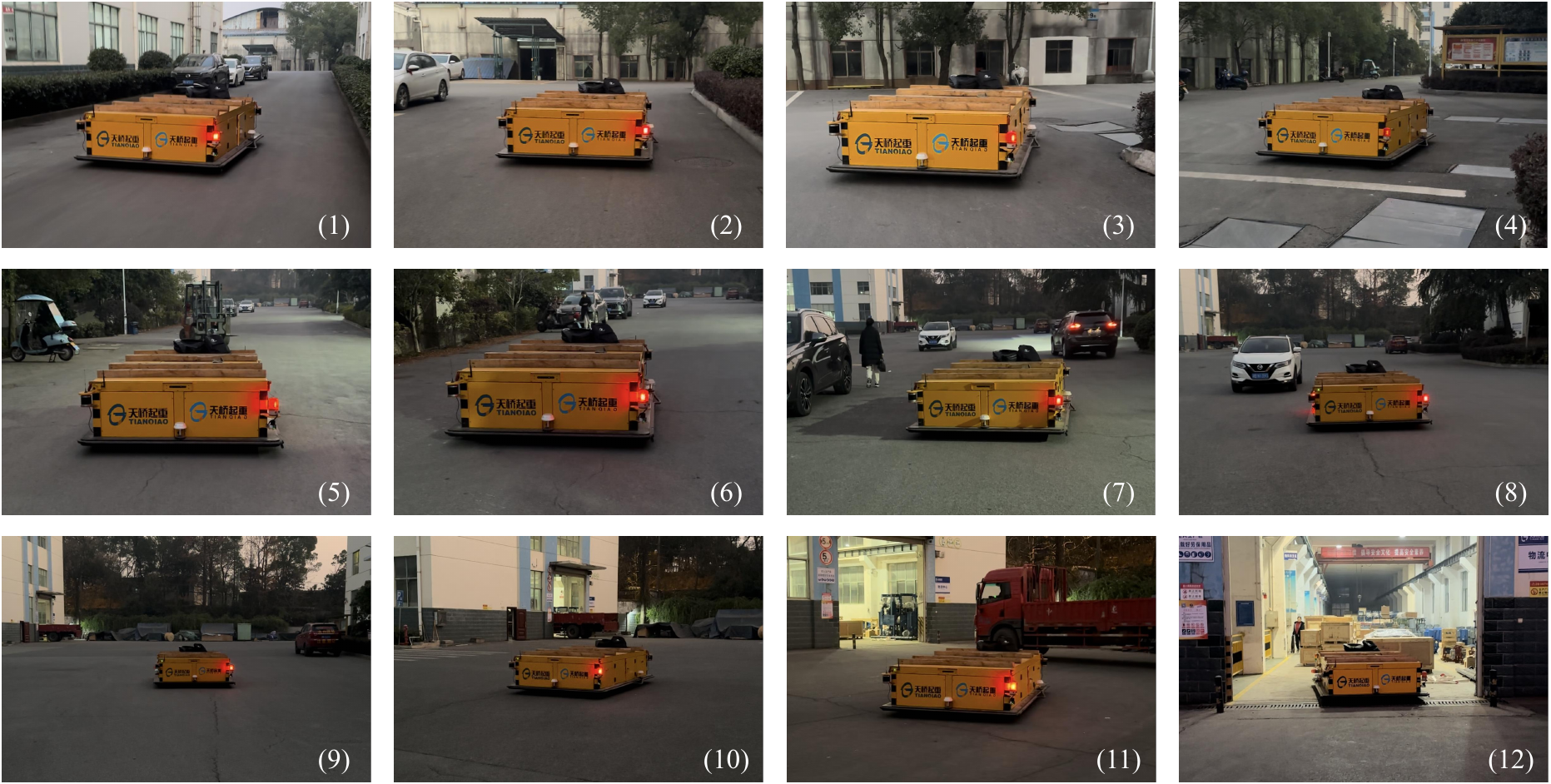}
\caption{Robot navigating an industrial factory setting, demonstrating real-time navigation with the hierarchical decision-making framework, successfully avoiding obstacles and reaching the destination.}
\label{fig:real_robot}
\end{figure}

To validate the real-world feasibility of the proposed hierarchical framework, we deployed the trained model onto a costom-designed 4WISD AMR with an Intel Core i7-8700T CPU and an NVIDIA RTX 1060 GPU. The field test was conducted in an active factory with predefined pathways, moving machinery, and human workers, evaluating the system’s adaptability to real-world challenges (Fig.~\ref{fig:real_robot}).

For long-distance navigation, the robot uses A* for global path planning, generating waypoints along the predefined route. At each step, the robot selects a waypoint and maintains a fixed distance, ensuring efficient navigation toward the final goal, even with dynamic obstacles. It continues from waypoint to waypoint until reaching the destination.

The trained model (3.5MB) was deployed without additional fine-tuning, demonstrating zero-shot transferability to real-world execution. The lightweight DRL network inference time is $1.43 \pm 11.33$ ms per step, and the fuzzy logic controller operates at $4.59 \pm 1.92$ ms per step. With a 10Hz control frequency, where all modules—hardware communication, map maintenance, and localization—must complete within 100 ms, the recorded computation times confirm the system meets real-time constraints.

\section{Conclusion and Future Work}
This paper introduced a hierarchical decision-making approach for 4WISD robots, combining a high-level DRL-based navigation policy with a low-level fuzzy logic controller to ensure kinematic feasibility. Experimental results demonstrate that the proposed framework outperforms conventional planners in navigation performance, achieving greater training efficiency and stability compared to a purely DRL-based approach. Future work will focus on exploring advanced reward-shaping techniques to enhance training efficiency, incorporating additional sensors to improve robustness in challenging environments, addressing transient misalignment and inter-wheel dragging issues during mode transitions, and extending the framework to coordinate multiple 4WISD robots for collaborative tasks in large-scale industrial applications.


\bibliographystyle{IEEEtran}
\bibliography{IEEEabrv,references}

\begin{thebibliography}{10}
\providecommand{\url}[1]{#1}
\csname url@rmstyle\endcsname
\providecommand{\newblock}{\relax}
\providecommand{\bibinfo}[2]{#2}
\providecommand\BIBentrySTDinterwordspacing{\spaceskip=0pt\relax}
\providecommand\BIBentryALTinterwordstretchfactor{4}
\providecommand\BIBentryALTinterwordspacing{\spaceskip=\fontdimen2\font plus
\BIBentryALTinterwordstretchfactor\fontdimen3\font minus
  \fontdimen4\font\relax}
\providecommand\BIBforeignlanguage[2]{{%
\expandafter\ifx\csname l@#1\endcsname\relax
\typeout{** WARNING: IEEEtran.bst: No hyphenation pattern has been}%
\typeout{** loaded for the language `#1'. Using the pattern for}%
\typeout{** the default language instead.}%
\else
\language=\csname l@#1\endcsname
\fi
#2}}

\bibitem{xiao2022motion}
X.~Xiao, B.~Liu, G.~Warnell, and P.~Stone, ``Motion planning and control for
  mobile robot navigation using machine learning: a survey,'' \emph{Autonomous
  Robots}, vol.~46, no.~5, pp. 569--597, 2022.

\bibitem{hang2021towards}
P.~Hang and X.~Chen, ``Towards autonomous driving: Review and perspectives on
  configuration and control of four-wheel independent drive/steering electric
  vehicles,'' in \emph{Actuators}, vol.~10, no.~8.\hskip 1em plus 0.5em minus
  0.4em\relax MDPI, 2021, p. 184.

\bibitem{cadena2016past}
C.~Cadena, L.~Carlone, H.~Carrillo, Y.~Latif, D.~Scaramuzza, J.~Neira, I.~Reid,
  and J.~J. Leonard, ``Past, present, and future of simultaneous localization
  and mapping: Toward the robust-perception age,'' \emph{IEEE Transactions on
  robotics}, vol.~32, no.~6, pp. 1309--1332, 2016.

\bibitem{zhu2021deep}
K.~Zhu and T.~Zhang, ``Deep reinforcement learning based mobile robot
  navigation: A review,'' \emph{Tsinghua Science and Technology}, vol.~26,
  no.~5, pp. 674--691, 2021.

\bibitem{lam2009omnidirectional}
T.~L. Lam, H.~Qian, and Y.~Xu, ``Omnidirectional steering interface and control
  for a four-wheel independent steering vehicle,'' \emph{IEEE/ASME transactions
  on mechatronics}, vol.~15, no.~3, pp. 329--338, 2009.

\bibitem{li2012model}
D.-Y. Li, Y.-D. Song, D.~Huang, and H.-N. Chen, ``Model-independent adaptive
  fault-tolerant output tracking control of 4ws4wd road vehicles,'' \emph{IEEE
  Transactions on Intelligent Transportation Systems}, vol.~14, no.~1, pp.
  169--179, 2012.

\bibitem{potluri2015path}
R.~Potluri and A.~K. Singh, ``Path-tracking control of an autonomous 4ws4wd
  electric vehicle using its natural feedback loops,'' \emph{IEEE Transactions
  on Control Systems Technology}, vol.~23, no.~5, pp. 2053--2062, 2015.

\bibitem{kosmidis2023novel}
A.~Kosmidis, G.~Ioannidis, G.~Vokas, and S.~Kaminaris, ``A novel real-time
  robust controller of a four-wheel independent steering system for ev using
  neural networks and fuzzy logic,'' \emph{Mathematics}, vol.~11, no.~21, p.
  4535, 2023.

\bibitem{setiawan2016path}
Y.~D. Setiawan, T.~H. Nguyen, P.~S. Pratama, H.~K. Kim, and S.~B. Kim, ``Path
  tracking controller design of four wheel independent steering automatic
  guided vehicle,'' \emph{International Journal of Control, Automation and
  Systems}, vol.~14, pp. 1550--1560, 2016.

\bibitem{liu2021nonlinear}
X.~Liu, G.~Wang, and K.~Chen, ``Nonlinear model predictive tracking control
  with c/gmres method for heavy-duty agvs,'' \emph{IEEE Transactions on
  Vehicular Technology}, vol.~70, no.~12, pp. 12\,567--12\,580, 2021.

\bibitem{ding2022trajectory}
T.~Ding, Y.~Zhang, G.~Ma, Z.~Cao, X.~Zhao, and B.~Tao, ``Trajectory tracking of
  redundantly actuated mobile robot by mpc velocity control under steering
  strategy constraint,'' \emph{Mechatronics}, vol.~84, p. 102779, 2022.

\bibitem{bae2023design}
B.~Bae and D.-H. Lee, ``Design of a four-wheel steering mobile robot platform
  and adaptive steering control for manual operation,'' \emph{Electronics},
  vol.~12, no.~16, p. 3511, 2023.

\bibitem{sprunk2017accurate}
C.~Sprunk, B.~Lau, P.~Pfaff, and W.~Burgard, ``An accurate and efficient
  navigation system for omnidirectional robots in industrial environments,''
  \emph{Autonomous Robots}, vol.~41, pp. 473--493, 2017.

\bibitem{shin2021model}
J.~Shin, D.~Kwak, and K.~Kwak, ``Model predictive path planning for an
  autonomous ground vehicle in rough terrain,'' \emph{International Journal of
  Control, Automation and Systems}, vol.~19, no.~6, pp. 2224--2237, 2021.

\bibitem{ma2022bi}
H.~Ma, F.~Meng, C.~Ye, J.~Wang, and M.~Q.-H. Meng, ``Bi-risk-rrt based
  efficient motion planning for autonomous ground vehicles,'' \emph{IEEE
  Transactions on Intelligent Vehicles}, vol.~7, no.~3, pp. 722--733, 2022.

\bibitem{yilmaz2022precise}
A.~Yilmaz, E.~Sumer, and H.~Temeltas, ``A precise scan matching based
  localization method for an autonomously guided vehicle in smart factories,''
  \emph{Robotics and Computer-Integrated Manufacturing}, vol.~75, p. 102302,
  2022.

\bibitem{zhu2017target}
Y.~Zhu, R.~Mottaghi, E.~Kolve, J.~J. Lim, A.~Gupta, L.~Fei-Fei, and A.~Farhadi,
  ``Target-driven visual navigation in indoor scenes using deep reinforcement
  learning,'' in \emph{2017 IEEE international conference on robotics and
  automation (ICRA)}.\hskip 1em plus 0.5em minus 0.4em\relax IEEE, 2017, pp.
  3357--3364.

\bibitem{tai2017virtual}
L.~Tai, G.~Paolo, and M.~Liu, ``Virtual-to-real deep reinforcement learning:
  Continuous control of mobile robots for mapless navigation,'' in \emph{2017
  IEEE/RSJ International Conference on Intelligent Robots and Systems
  (IROS)}.\hskip 1em plus 0.5em minus 0.4em\relax IEEE, 2017, pp. 31--36.

\bibitem{fan2018crowdmove}
T.~Fan, X.~Cheng, J.~Pan, D.~Manocha, and R.~Yang, ``Crowdmove: Autonomous
  mapless navigation in crowded scenarios,'' \emph{arXiv preprint
  arXiv:1807.07870}, 2018.

\bibitem{jang2021hindsight}
Y.~Jang, J.~Baek, and S.~Han, ``Hindsight intermediate targets for mapless
  navigation with deep reinforcement learning,'' \emph{IEEE Transactions on
  Industrial Electronics}, vol.~69, no.~11, pp. 11\,816--11\,825, 2021.

\bibitem{zhu2022hierarchical}
W.~Zhu and M.~Hayashibe, ``A hierarchical deep reinforcement learning framework
  with high efficiency and generalization for fast and safe navigation,''
  \emph{IEEE Transactions on Industrial Electronics}, vol.~70, no.~5, pp.
  4962--4971, 2022.

\bibitem{miranda2023generalization}
V.~R. Miranda, A.~A. Neto, G.~M. Freitas, and L.~A. Mozelli, ``Generalization
  in deep reinforcement learning for robotic navigation by reward shaping,''
  \emph{IEEE Transactions on Industrial Electronics}, 2023.

\bibitem{guo2023optimal}
H.~Guo, Z.~Ren, J.~Lai, Z.~Wu, and S.~Xie, ``Optimal navigation for agvs: A
  soft actor--critic-based reinforcement learning approach with composite
  auxiliary rewards,'' \emph{Engineering Applications of Artificial
  Intelligence}, vol. 124, p. 106613, 2023.

\bibitem{wang2024heuristic}
Y.~Wang, Y.~Xie, D.~Xu, J.~Shi, S.~Fang, and W.~Gui, ``Heuristic dense reward
  shaping for learning-based map-free navigation of industrial automatic mobile
  robots,'' \emph{ISA transactions}, vol. 156, pp. 579--596, 2025.

\bibitem{haarnoja2018soft}
T.~Haarnoja, A.~Zhou, K.~Hartikainen, G.~Tucker, S.~Ha, J.~Tan, V.~Kumar,
  H.~Zhu, A.~Gupta, P.~Abbeel, \emph{et~al.}, ``Soft actor-critic algorithms
  and applications,'' \emph{arXiv preprint arXiv:1812.05905}, 2018.

\bibitem{haarnoja2017reinforcement}
T.~Haarnoja, H.~Tang, P.~Abbeel, and S.~Levine, ``Reinforcement learning with
  deep energy-based policies,'' in \emph{International conference on machine
  learning}.\hskip 1em plus 0.5em minus 0.4em\relax PMLR, 2017, pp. 1352--1361.

\bibitem{fox1997dynamic}
D.~Fox, W.~Burgard, and S.~Thrun, ``The dynamic window approach to collision
  avoidance,'' \emph{IEEE Robotics \& Automation Magazine}, vol.~4, no.~1, pp.
  23--33, 1997.

\bibitem{rosmann2015timed}
C.~R{\"o}smann, F.~Hoffmann, and T.~Bertram, ``Timed-elastic-bands for
  time-optimal point-to-point nonlinear model predictive control,'' in
  \emph{2015 european control conference (ECC)}.\hskip 1em plus 0.5em minus
  0.4em\relax IEEE, 2015, pp. 3352--3357.

\end{thebibliography}

\addtolength{\textheight}{-12cm}   


\end{document}